\documentclass[10pt,twocolumn,letterpaper]{article}

\usepackage{iccv}
\usepackage{times}
\usepackage{epsfig}
\usepackage{graphicx}
\usepackage{amsmath}
\usepackage{amssymb}
\usepackage{multirow, booktabs}
\usepackage{float}
\usepackage[pagebackref=true,breaklinks=true,letterpaper=true,colorlinks,bookmarks=false]{hyperref}
\usepackage[capitalize]{cleveref}
\crefname{section}{Sec.}{Secs.}
\Crefname{section}{Section}{Sections}
\Crefname{table}{Table}{Tables}
\crefname{table}{Tab.}{Tabs.}
\usepackage{multirow}
\usepackage{caption2}
\usepackage{authblk}

\iccvfinalcopy % *** Uncomment this line for the final submission

\def\iccvPaperID{1975} % *** Enter the ICCV Paper ID here

\ificcvfinal\fi

\makeatletter
\renewcommand{\@fnsymbol}[1]{}
\makeatother

\title{FBLNet: FeedBack Loop Network for Driver Attention Prediction}

\author{
  Yilong Chen,
  Zhixiong Nan$^*$\thanks{Zhixiong Nan$^*$ is the corresponding author.},
  Tao Xiang
    \thanks{This work was supported by the National Natural Science Foundation of China under Grant (62006180 and 62106026) and CCF-AFSG Research Fund (RF20220009, 02160026050171).}\\
  \textsuperscript{}Chongqing University, Chongqing, China \\
}
\begin{document}
\maketitle

%%%%%%%%% ABSTRACT
\begin{abstract}
The problem of predicting driver attention from the driving perspective is gaining increasing research focus due to its remarkable significance for autonomous driving and assisted driving systems. 
The driving experience is extremely important for safe driving, 
a skilled driver is able to effortlessly 
predict oncoming danger (before it becomes salient) based on the driving experience and quickly pay attention to the corresponding zones.
% especially when threatening traffic participants have not become salient. 
However, the nonobjective driving experience is difficult to model, so a mechanism simulating the driver experience accumulation procedure is absent in existing methods, and the current methods usually follow the technique line of saliency prediction methods to predict driver attention.  
% so we have not found a work that simulates human experience accumulation procedure to assist driver attention prediction,
In this paper, we propose a FeedBack Loop Network (FBLNet), which attempts to model the driving experience accumulation procedure. By over-and-over iterations, FBLNet generates the incremental knowledge that carries rich historically-accumulative and long-term temporal information. The incremental knowledge in our model is like the driving experience of humans.
Under the guidance of the incremental knowledge, our model fuses the CNN feature and Transformer feature that are extracted from the input image to predict driver attention. 
Our model exhibits a solid advantage over existing methods, achieving an outstanding performance improvement on two driver attention benchmark datasets. 
\end{abstract}

\begin{figure}[t]
    \setlength{\abovecaptionskip}{0.cm}
    \setlength{\belowcaptionskip}{-0.5cm}
	\centering
	\includegraphics[width=0.49\textwidth]{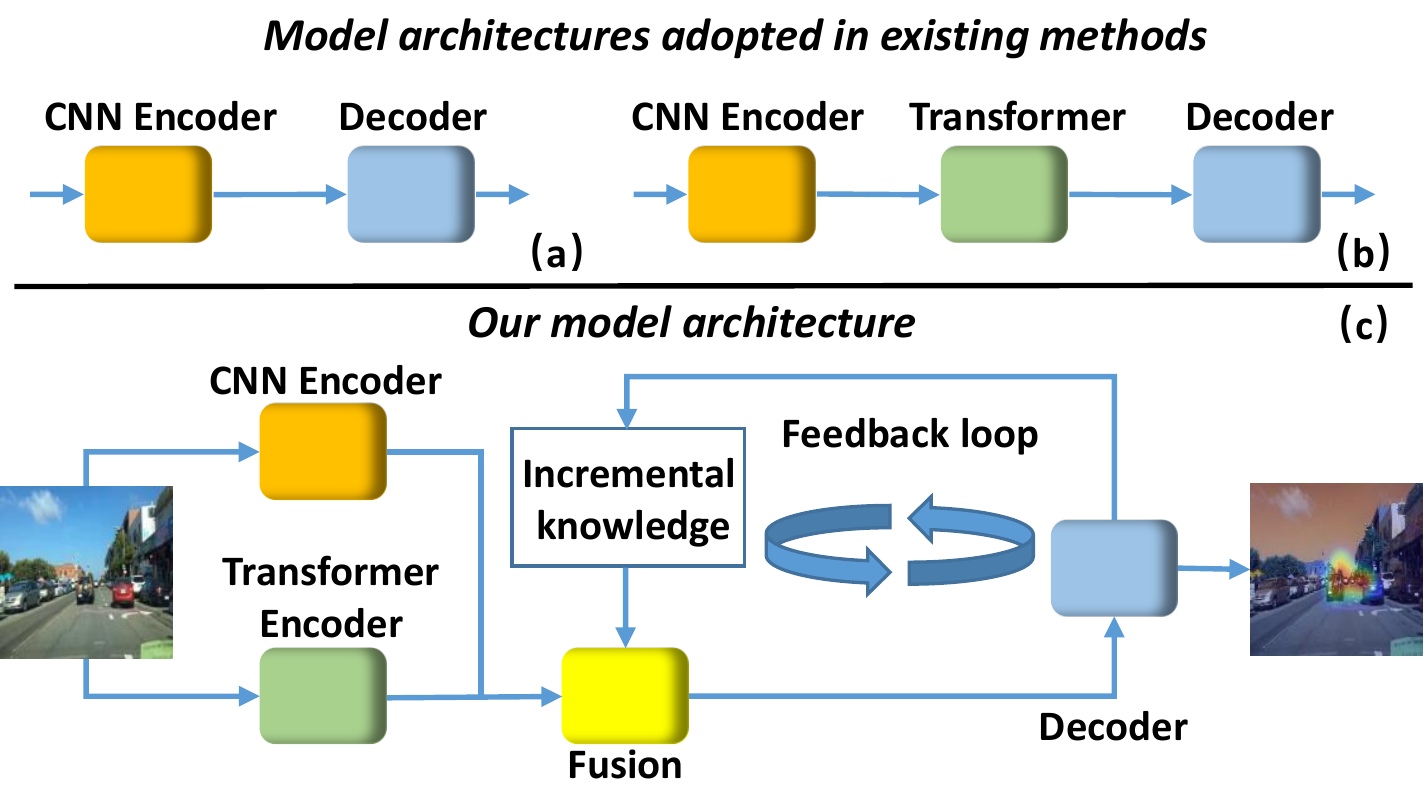}
	\caption{The comparison of the mainstream existing architectures (i.e., \textbf{(a)} CNN based encoder-decoder architecture and \textbf{(b)} CNN-
Transformer based encoder-decoder architecture) and our architecture (i.e., \textbf{(c)} FBLNet).
	}
	\label{fig:introduction}
\end{figure}

%%%%%%%%% BODY TEXT
\section{Introduction}
% Driving is an indispensable part of our daily life which has the characteristics of large number of participants, long time and wide distribution. But even more remarkable is the high risk of driving. 
Accurately predicting where a driver should look is extremely important for assisted driving\cite{startADAS}, accident avoidance\cite{startTraffic}, and driverless driving\cite{driveless}. According to the road traffic injuries report $\footnote{https://www.who.int/zh/news-room/fact-sheets/detail/road-traffic-injuries}$ of WHO in 2022, nearly 1.3 million people die in traffic accidents every year, and a high proportion of death result from driver attention distraction.
%Among these accidents,  According to the report (a news need to add)--------------
%in the UN News$\footnote{https://news.un.org/zh/story/2010/05/131362}$, distracted driving causes about 1.6 million crashes and 6,000 deaths in the United States each year, accounting for 16 percent of all crash deaths. 
Many accidents could be avoided if a system is able to accurately predict where a driver should look and timely send the attention distraction warning. Therefore, driver attention prediction is significant for developing safer autonomous driving or assisted driving systems. 
% appropriate warning for driver's distraction. 
%Accurately predicting where a driver should be looking at and appropriate warning for his distraction can avoid many accidents. 
On account of the remarkable research significance of driver attention prediction, it is attracting increasingly wide research focuses, and many excellent models \cite{rainy,LiQiang2022,bdda,palazzi2017learning} have been proposed in recent years. 

The traffic scene is challenge and complex. It is highly dynamic, random and diverse. A driver may confront various scenarios, such as cutting in cars, crossing pedestrians and emergently-appearing motorbikes. 
To safely drive in such scenarios, \textbf{driving experience} is the key point. A skilled driver could effortlessly predict the most likely area where the danger may appear. For example, when a side front car is following another slowly driving car in front of it, the side front car may change to our lane, which is dangerous and we should pay more attention to it to avoid the collision. What's more, an experienced driver could predict the potentially dangerous zones according to the current traffic scene even if the danger is not visible at the current time, and pay attention to the dangerous zones to guarantee safety. Therefore, it has great significance if we could design a driver attention prediction model that possesses the ability to accumulate the driving experience like a human.

 Unfortunately, after reviewing the recent literature regarding driver attention prediction, we have not found a work that is struggling in this direction. Existing models can be coarsely divided into two categories. One kind of model is CNN (Convolution Neural Network) based encoder-decoder, as shown in Fig.~\ref{fig:introduction}(a). The other kind is CNN-Transformer based encoder-decoder, as shown in Fig.~\ref{fig:introduction}(b), which first uses CNN to extract local features, then adopts Transformer to explore the global context of the local features, and finally adds a decoder at the end of Transformer. These models have remarkably driven the development of this field. However, their core idea is similar to that of models for saliency prediction, ignoring the importance of \textit{long-term historically-accumulative information (similar to the driving experience)} for driver attention prediction. As a result, driver attention can be hardly predicted when saliency target is not consistent with driver attention.

Aiming at designing a driver attention prediction model that is able to simulate human driving experience accumulation procedure, this paper proposes a \textbf{F}eed\textbf{B}ack \textbf{L}oop \textbf{Net}work ({FBLNet}). Essentially different from existing models, the novelty of {FBLNet} is the feedback loop mechanism that continuously fetches data from the decoder module in every round of training procedure to iteratively update the \textbf{incremental knowledge}. The feedback loop mechanism enables the {incremental knowledge} to carry historically-accumulative and long-term temporal information, making the whole model have a similar ability as a human accumulating the driving experience.

Fig.~\ref{fig:introduction}(c) illustrates the coarse structure of our model. Given an input image, CNN encoder and Transformer encoder extract the features of the input image, and the feedback loop is responsible for updating the incremental knowledge. The fusion module accepts image features and the incremental knowledge as the input, and outputs the fusion feature. The fusion feature is upsampled by the decoder module to construct the driver attention heatmap. To evaluate our model, a series of comparison and diagnostic experiments are conducted on two public datasets. The comparison experiment results show that our model has a solid advantage over the baselines, achieving significant improvement on six metrics. The diagnostic experiment results validate the effectiveness of our proposed feedback loop module, fusion module and encoder module. 

% Based on the above ignore, in this work, we take this part of forgetting into account and propose a model named FBLNet. As can be seen in Fig.~\ref{fig:introduction}, the major novelty of our FBLNet is the FeedBack Loop and a fusion model. Through the operation of the above two structures, we can feedback the incremental knowledge which is a supplement to the blank of driving experience to the fusion model through the FBL structure in every training. The incremental knowledge , updates itself in each round of training and finally approaches the prior knowledge of a driver. For the fusion, it fuse the local information and global information from CNN-Transformer Encoder and incremental knowledge together. To prove the rationality of our model, we design a series of quantitative experiments to verify the performance of our model, and design relevant ablation experiments to verify the significance of our CNN-Transformer Encoder, FBL structure and fusion model. Finally, excellent results are obtained.

The contributions of this paper are as follows. 
\textbf{1)} As Fig~\ref{fig:introduction} shown, our proposed FBLNet is essentially different from all existing methods for driver attention prediction. Our FBLNet can utillze the historically-accumulative and long-term temporal information by a feedback loop structure like a human driver accumulating his driving experience. Extensive experimental results on two driver attention benchmark datasets show that FBLNet outperforms existing models for attention prediction .
%The experiment results validate that the FBLNet is a tiny but successful attempt to make a model that has the human-like ability to model the nonobjective concept such as experience and knowledge. 
%\textbf{2)} The feedback loop mechanism of FBLNet is an extensible technique that could be generalized to solve other problems, and it can be easily applied to other networks. 
\textbf{2)}  We also propose the incremental knowledge guided fusion module. 
%好处
This novel fusion module can explore the inter-relationship between different kinds of features and  guides the CNN feature and the Transformer feature to update in accordance with incremental knowledge during the training process. 
%This fusion network not only strengthens the CNN feature and Transformer feature under the guidance of the incremental knowledge, but also models the relations among different kinds of features via the Transformer idea.
% The proposed structure-aware Transformer approach has been properly analyzed and verified through extensive experiments over DR(Eve)VY\cite{dr(eve)vy}, BDDA\cite{bdda} and DADA\cite{dada} datasets to validate its potential in solving driver attention prediction problem.
%------------------------------------------------------------------------
\section{Related Works}
%Driver attention prediction aims at predicting the driver attention area by analyzing the images from the driver's perspective. Driver attention prediction is actually trying to realize a model to observe the driving scene like the driver. 
%Accurately predicting where a driver should look is extremely important for assisted driving\cite{startADAS}, accident avoidance\cite{startTraffic}, and driverless driving \cite{startAutonomous}. 

Methods for this problem can be roughly divided into three types: \textbf{1)} Conventional methods, \textbf{2)} Convolutional neural network methods and \textbf{3)} Transformer methods.
%In the early days before convolutional neural networks, the dominant method is to use the algorithm that simulates the driver's attention behavior to predict. Later, with the continuous development of convolutional neural network in the field of computer vision, the use of convolution operation has become the first choice of researchers. This method has well extracted the features in the driving image and analyzed the relationship between pixels, which has greatly promoted the study of this problem. At present, the successful application of Transformer in the NLP field makes more and more scholars apply Transformer in the CV field. The proposal of VIT makes this idea possible. Besides, Transformer's excellent computing power, good expansibility and extensibility indicate its huge potential in this problem.
\subsection{Conventional Methods}
In the early days, the dominant method mainly adopts the probability model to predict attention. Itti \textit{et al.}\cite{1998Itti,2004Itti} proposes a feature map combining image color, attributes and orientation features, which uses a dynamic neural network to select the concerned positions according to the significance. At that time, the mainstream is using a mathematical algorithm, to simulate human attention mechanisms, such as the Borji graph model based on probabilistic inference and top-down research methods\cite{2014Borji}, which uses the Bayesian network. Bayesian network is also used in \cite{pang2008}, which uses a dynamic Bayesian model and combines visual salience with people's cognitive state to predict. In the traffic significance detection model\cite{2016Deng}, a bottom-up and top-down combined model framework is constructed by using road vanishing points as guidance. Ban \textit{et al.}\cite{Ban2010} also uses the top-down and bottom-up fusion model, the core of his top-down approach is adaptive resonance theory (ART), which is also used in \cite{fang2003}. \cite{Heracles2009} is inspired by the biological concept of the ventral attention system, which focuses on incidents of low significance but high stimulation. These above methods do not use CNN, but their top-down and bottom-up ideas inspire later researches.
\subsection{Convolutional Methods}
With the continuous development of the convolutional neural network in the field of computer vision, CNN-based methods have become the first choice for researchers. 
These methods can be divided into two types, bottom-up type and a combination of bottom-up and top-down type.
In the bottom-up approach, \cite{2017Tawari} uses Bayesian modeling, and adds a fully convolutional neural network to process the image. Palazzi \textit{et al.}\cite{2019Palazzi, palazzi2017learning} uses C3D network. \cite{Hammer} also uses C3D convolution operation. Lattef \textit{et al.}\cite{2022Gan} uses a generative model which has a simple structure. Deng \textit{et al.}\cite{rainy} proposes a Dual-Branch model that divides the input into a single frame and continuous frames. 
%The single frame enters CNN and the continuous frames enter LSTM. 
\cite{U2-Net,pang2020multi,SimpleNet} are all based on U-Net.
%, and make improvement. 
Meanwhile, many scholars integrate some top-down research methods into the convolutional neural network\cite{redundant,bdda,LiQiang2022}. 
Fang \textit{et al.}\cite{dada} proposes SCAFNet network is a method of simultaneous processing and a fusion of source images and semantic images.
And a big step forward for the top-down method is the application of attention mechanism\cite{2021Deng,wang2020convlstm}. 
%ADA uses generic decoder with attention mechanism to replicate top-down mechanism and bottom-up mechanism respectively\cite{ADA}. DSTANet is a fusion of ConvLSTM and attention gate\cite{2021Deng}, a method that introduces information about driving space-time and simulates human attention mechanisms. 
%ConvLSTM is also used in the research of \cite{wang2020convlstm} and \cite{LiQiang2022}. The former uses three attention mechanisms to analyze the features, and the latter integrated ConvLSTM into 3D-CNN. 
%\cite{bdda} starts from human vision, looking for the important frame that is a significant change frame in multiple frames.  And \cite{redundant} further considers the redundant regions by using a convolutional method. 
MEDRIL\cite{baee2021medirl} is the method of maximum entropy depth reverse reinforcement learning.
%to predict attention for accident-prone situations. 
%These methods only use the information from local detailed features, but can not effectively capture global information that is also important for driver attention.
These methods solely rely on local detailed features, thus failing to effectively capture the crucial global information necessary for accurate driver attention prediction.
\subsection{Transformer Methods}
At present, the successful applications of Transformer in the NLP field impels scholars to apply Transformer in the CV field and many excellent results are achieved. The applications of Transformer in the field of computer vision can be divided into: \textbf{1)} pure use of Transformer, \textbf{2)} integration of Transformer with CNN. \cite{MA2022PureTrans} is a pure Transformer study. 
Also, many researchers notice the different characteristics between CNN and Transformer,  meanwhile, plenty of excellent research has been done to explore it. \cite{2022swinT,PGNet} uses U-Net and integrates the Swin-transformer. \cite{Huang2022Tans} fuses Transformer with a convolution network, then inputs the results into a ConvLSTM for time information learning. 
%These researches suggest that Transformer can also be used in driver attention prediction. But, only focusing on the driving scene's local features and global features are not enough to accurately predict driver attention. Historical driving experience also influences where drivers focus, which is not considered by former researchers.
These studies indicate the potential of using Transformers in driver attention prediction. However, solely relying on the local and global features of the driving scene is insufficient for precise driver attention prediction. Neglecting the influence of historical driving experience, as previous researchers have done, can lead to incomplete results. Taking into account the driver's accumulated experience is crucial for a more accurate and comprehensive understanding of where their focus lies during driving tasks.
\begin{figure*}[h]
	\centering
	\setlength{\abovecaptionskip}{0.cm}
	\setlength{\belowcaptionskip}{-0.3cm}
	\includegraphics[width=1\textwidth]{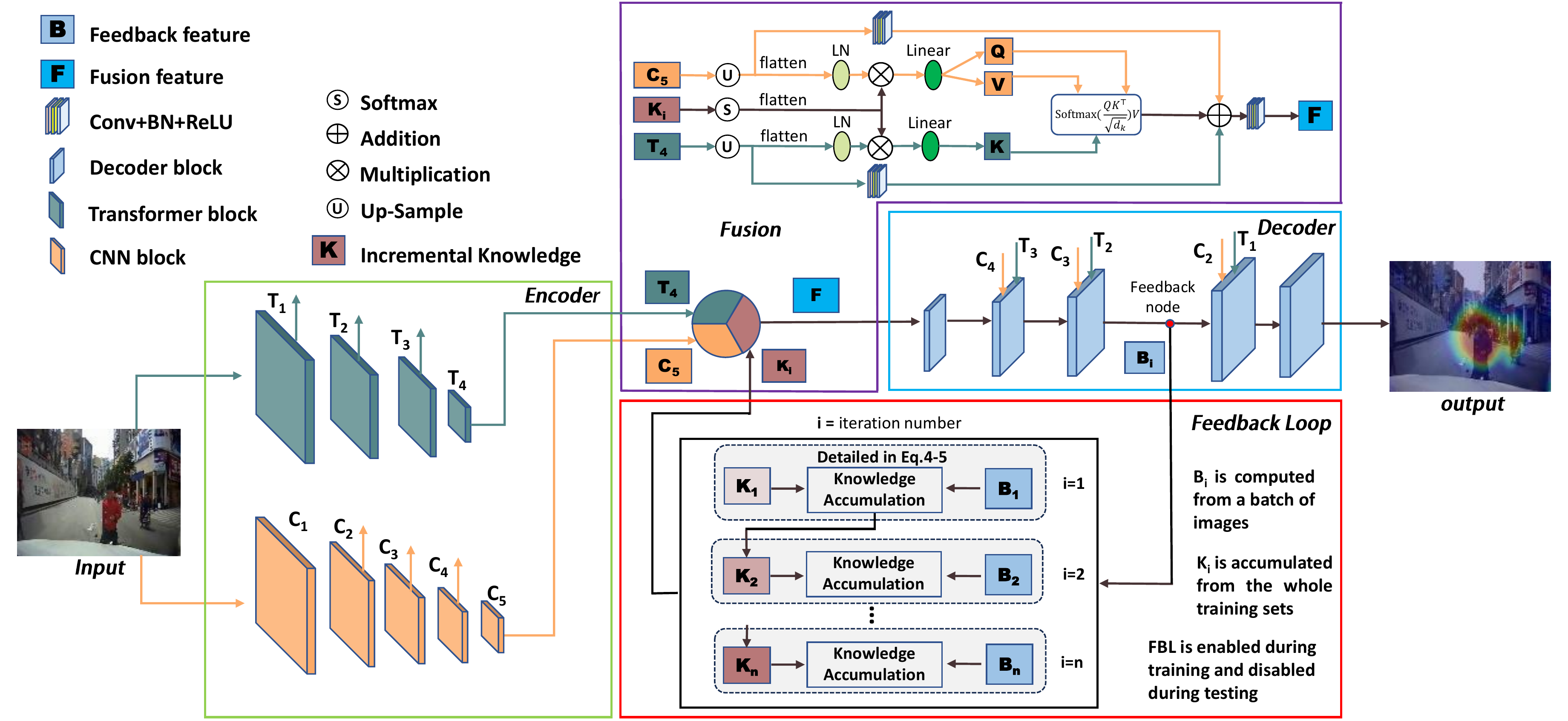}
	\caption{The overview of our method. \textbf{Encoder} module extracts CNN feature $\textbf{C}_5$ and Transformer feature $\textbf{T}_4$ from input image. \textbf{Feedback Loop} module continuously accepts the feedback feature $\textbf{d}_i$ from the Decoder module, and iteratively generates the incremental knowledge $\textbf{K}_i$. \textbf{Fusion} module fuses the $\textbf{C}_5$, $\textbf{T}_4$ and $\textbf{K}_i$ to generate the fusion feature $\textbf{F}$. \textbf{Decoder} module takes $\textbf{F}$ as input, and finally outputs the driver attention prediction.}
	\label{main}
\end{figure*}
\section{Approach}
\subsection{Overview}
The goal of this work is to predict driver attention $\textbf{A}$ given the image $\textbf{I}$ captured from the perspective of drivers, which is formulated as:
%$ \mathbb{R}^{3\times W\times H}  \to \left \{ 0,1 \right \} ^{W\times H}$. 
\begin{equation}
\textbf{A} = \mathcal{N}(\textbf{I})
\end{equation}
where $\textbf{I} \in \mathbb{R}^{3\times W\times H}$, $ \textbf{A} \in (0,1)^{W\times H}$ and $\mathcal{N}$ represents a driver attention prediction network.

Different from the prior conventional encoder-decoder network structure we propose the FeedBack Loop (FBL) encoder-decoder network, which is called FBLNet. The overview of FBLNet is illustrated in Fig.~\ref{main}.  FBLNet is composed of four modules, including CNN-Transformer Feature Extraction (\S ~\ref{section_feature_extraction}), FeedBack Loop (\S ~\ref{section_Feedback Loop}), Incremental Knowledge Guided Fusion (\S ~\ref{section_Fusion Model}) and Decoder (\S ~\ref{section_decoder}). The CNN-Transformer Feature Extraction module extracts the CNN feature $\textbf{C}$ and the Transformer feature $\textbf{T}$ from $\textbf{I}$. FBL module continuously accepts the feedback information $\textbf{d}$ from the Decoder module, and iteratively generates the incremental knowledge $\textbf{K}$. In the Incremental Knowledge Guided Fusion module, $\textbf{C}$, $\textbf{T}$, and $\textbf{K}$ are fused to produce the fusion feature $\textbf{F}$. The Decoder module takes $\textbf{F}$ as input, and finally outputs $\textbf{A}$. 

% In this work, we design a Transformer network architecture based on feedback loop. Its feature extraction part is composed of the Encoder part of CNN and Transformer. The two features obtained are conducted by a feedback loop structure, and the feedback loop provides a Historical Global Information. The Historical Global Information, convolution results and Transformer results are simultaneously fused in a fusion module, and the output, fusion feature comes into Decoder model. The Decoder provides feedback feature and final output, the former return to loop and becomes the Historical Global Information that acts on fusion model next round, the latter is our model's prediction. Therefore, the whole architecture is mainly composed of four main components: Encoder, Decoder, fusion model and Feedback Loop. Two encoder structures for feature extraction, which focus on the local features and global features of the image respectively. However, if there are two input segments, a fusion module is definitely needed to integrate the information of these two parts. Therefore, we designed an information fusion module to fuse the result lines of the two decoder structures, and the fusion process is influenced by the cyclic network at the same time. The fused result is then passed to a single-stream decoder module, which outputs the final prediction. Finally, the feedback loop which is the most important structure, pass the feedback feature to the fusion module, so that the network can better combine the global information and local information.

\subsection{CNN-Transformer Feature Extraction}
\label{section_feature_extraction}
%mo
%Different from other scenes,
Traffic scenes are extremely dynamic, complex and random. In addition, traffic participants are numerous and different participants always have cooperation and competitive  relationships. Considering the specific characteristics of traffic scenes, to accurately predict the driver's attention, it is essential to simultaneously encode the local information (e.g., cutting in cars and persons) and analyze the global information (e.g., multiple participants at the crossroads). Therefore, different from most existing methods, we design the CNN-Transformer Encoder that consists of two pathways, one is the CNN pathway and the other is the Transformer pathway. The former enables the encoder to extract the local detailed feature, while the latter allows to extract the global long-range dependency feature.

% In general, our Encoder consists of two parts, a CNN and a Transformer, which are two main methods for feature extraction.  The former focuses on local information, while the latter has a good ability to capture global information. For our problem of driver's attention prediction, it contains not only pixel-level information of the image itself, but also various connections between different areas in a picture. Therefore, it is necessary to understand the local information of the visual field from the driver's perspective and analyze the global information from a global perspective. In this way, it is possible to accurately predict where the driver should look. Therefore, the feature extraction method of CNN and Transformer are used in our work.

%CNN
The feature extraction of the Convolution Neural Network pathway is formulated as:
\begin{equation}
%\{\textbf{C}_{i}|i=1,2,3,4,5\} = {\mathcal{N}_{cnn}}(I)\\
\left \{\textbf{C}_{i}  \right \}_{i=1}^{5} = \mathcal{N}_{cnn}(\textbf{I}) 
\label{eq:cnn feature}
\end{equation}
where $\textbf{I}$ is the input image with the size of ${3}\times{224}\times{224}$, and 
 $\mathcal{N}_{cnn}$ denotes the backbone of CNN. Five CNN features $\{\textbf{C}_{i}\}_{i=1}^{5}$ are extracted, and the size of $\textbf{C}_{i}$ is $\left( {{64} \times {2^{{\rm{i}} - {\rm{1}}}}} \right) \times  \frac{{224}}{{{2^{\rm{i}}}}} \times \frac{{224}}{{{2^{\rm{i}}}}} $. Among them, $\{\textbf{C}_{i}\}_{i=2}^{4}$ will be fused with decoder features to strengthen expressions (see \S ~\ref{section_decoder}), and  $\textbf{C}_{5}$ will be used as one input of fusion module (see \S ~\ref{section_Fusion Model}).

%Trans
The feature extraction of the Transformer pathway is expressed as:
\begin{equation}
%\{\textbf{T}_{i}|i=1,2,3,4\} = {\mathcal{N}_{Trans}}(I)
\left \{\textbf{T}_{i}  \right \}_{i=1}^{4} = \mathcal{N}_{Trans}(\textbf{I}) 
\label{eq:transformer feature}
\end{equation}
where $\mathcal{N}_{Trans}$ denotes the Transformer blocks, generating four Transformer features $\{\textbf{T}_{i}\}_{i=1}^{4}$, and the size of $\textbf{T}_{i}$ is $\left( {64 \times {2^{\rm{i} - {\rm{1}}}}} \right) \times \frac{{{\rm{56}}}}{{{2^{{\rm{i}} - {\rm{1}}}}}} \times \frac{{{\rm{56}}}}{{{2^{{\rm{i}} - {\rm{1}}}}}} $.  $\{\textbf{T}_{i}\}_{i=1}^{3}$ will be fused with decoder features to strengthen expressions (see \S ~\ref{section_decoder}), and $\textbf{T}_{4}$ will be used as one input of fusion module (see \S ~\ref{section_Fusion Model}).
\subsection{Feedback Loop}
\label{section_Feedback Loop}
%How to better use the information captured by encoder into prediction? Different from the current research of other scholars, our work proposed a Feedback Loop structure. 
The major novelty of our FBLNet is the FeedBack Loop (FBL) structure, which is inspired by the core theory (i.e., feedback) of Wiener's cybernetics claiming that the feedback mechanism makes a system stable by conveying information from the back end to the front end. As shown in Fig.~\ref{main}, FBL continuously fetches the decoder feature from a feedback node and iteratively carries the feedback feature forward to accumulate it to the incremental knowledge.
It is important to note that FBL is exclusively employed during the training phase. Once the training phase is completed, $\textbf{K}$ remains constant. This process is analogous to a driver accumulating driving experience from various driving scenarios (training set), and the accumulated experience ($\textbf{K}$) can be effectively utilized when the driver encounters new road conditions (testing set).
%Every feature which is conveyed to the front can be regarded as a predictive behavior of a model in one iteration.  We accumulate this predictive behavior, transfer it to another form of feature which we call incremental knowledge, and update it in each iteration. This process is similar to human learning driving, and this kind of knowledge can be utilized in the next round of prediction.

%The incremental knowledge is a feature map with all elements initialized as ``1''. The feedback loop is repeated in each forward propagation step, thus the Incremental Knowledge is constantly updated. This procedure is formulated as: 
%The incremental knowledge is represented as a feature map with all elements initialized to ``1". During each forward propagation step, the feedback loop is repeated, leading to continuous updates of the Incremental Knowledge. This iterative process can be formulated as follows:
The incremental knowledge is physically represented as a feature map, where all elements are initialized to ``1". The feedback loop is repeated in each forward propagation step, thus the Incremental Knowledge is constantly updated. This procedure is formulated as: 
\begin{equation} 
{\textbf{K}_i} = \left\{ \begin{aligned}
&1,&i = 1 \\
&\mathcal{F}_{iter}({\textbf{K}_{i - 1}},{\textbf{B}_{i-1}}), &i \ge 2
\end{aligned}
\right
.
\label{eq:K_i}
\setlength{\belowdisplayskip}{8pt}
\setlength{\abovedisplayskip}{8pt}
\end{equation}
where $i$ represents the iteration step number, $\textbf{K}_1$ is the initialization matrix which has the same size as the feedback feature, $\textbf{K}_i$ is the incremental knowledge after $i$ iterations, and we note that each iteration equals that a batch of data goes through a complete forward propagation procedure. 
${F}_{iter}$ in Eq.~\ref{eq:K_i} is the iteration rule defined as:
\begin{equation}
\footnotesize
\mathcal{F}_{iter}({\textbf{K}_{i-1}},\textbf{B}_{i-1}) = {\mathop{\rm mean}\nolimits} ({\mathop{\rm ReLU}\nolimits} ({\mathop{\rm BN}\nolimits} ({\mathop{\rm Conv}\nolimits} (\textbf{K}_{i-1} \oplus\textbf{B}_{i-1}))) + \textbf{K}_{i-1})
\label{eq:K_i_iter}
\setlength{\belowdisplayskip}{8pt}
\setlength{\abovedisplayskip}{8pt}
\end{equation}
$\textbf{B}_{i-1}$ in Eq.~\ref{eq:K_i} is the feedback feature $\textbf{B}$ at ${(i-1)}_{th}$ iteration, and $\oplus$ is represented for concatenation. $\textbf{B}$ is fetched from a certain feedback node in decoder:
\begin{equation}
    \textbf{B}=\textbf{d}_j, \textbf{d}_j\in\{ \textbf{d}_j\}_{j=0}^{4}
\label{eq:feedback_feature}
\setlength{\belowdisplayskip}{10pt}
\setlength{\abovedisplayskip}{10pt}
\end{equation}
where $\textbf{d}_j$ is the decoder feature after $j$ decoder layers. 
%In our proposed FBLNet, the number of layers of the decoder is 5.

We can find that $\textbf{B}$ could be fetched from any decoder layers, thus ``where to start feedback" is a crucial section. 
According to the theory in classic Wiener's cybernetics, the feedback node is near to the output, which impels to select $\textbf{B} = \textbf{d}_4$. However, considering the characteristics of the deep neural network, we selected $\textbf{B} = \textbf{d}_2$, and the analysis will be detailed in the diagnostic experiment (\S ~\ref{subsection:Diagnostic Experiment}).

% \noindent{\textit{\textbf{Where to start feedback}}}. As shown in Fig.~\ref{main}, we need to choose a node from Decoder part to become our feedback node, which is equivalent to choosing a $\textbf{d}_i$ to become our $\textbf{B}$: 
As a short summary, the core idea of our proposed FeedBack Loop is to feedback the decoder feature to iteratively update the incremental knowledge. By continuous feedback, the information in the temporal dimension is increasingly accumulated to the incremental knowledge, making it carry rich historically-accumulative and long-temporal information, thus the incremental knowledge could perform a crucial guiding role for driver attention prediction. 

\subsection{Incremental Knowledge Guided Fusion}
\label{section_Fusion Model}
The input of our fusion module is CNN feature $\textbf{C}_5$ in Eq.~\ref{eq:cnn feature}, Transformer feature $\textbf{T}_4$ in Eq.~\ref{eq:transformer feature}, and the incremental knowledge $\textbf{K}_i$ in Eq.~\ref{eq:K_i}. The output is the fusion feature denoted as $\textbf{F}$. Different from the classic fusion mechanisms (e.g., addition and concatenation), we propose an incremental knowledge guided and Transformer based fusion mechanism. $\textbf{K}_i$ is used to guide the transition of $\textbf{C}_5$ and $\textbf{T}_4$, and Transformer is adapted to allow the network to learn the potential inter-relationship between different kinds of features.

% Transformer based fusion method to model the guiding impact of $\textbf{K}_i$ on $\textbf{C}_5$ and $\textbf{T}_4$, 

% and the residual idea is utilized to reserve the original information in $\textbf{C}_5$ and $\textbf{T}_4$. 
% We have obtained the local information extracted by CNN($\textbf{C}_5$), the global information from Transformer($\textbf{T}_4$), and the incremental knowledge($\textbf{K}_i$) from feedback loop. How to effectively integrate these three information has become the core problem of our work. To better fuse these information as fusion feature($\textbf{B}$), we design a fusion model:
%In this paper, the Decoder intermediate output of each prediction process can be considered as a probability graph describing whether the position of each pixel is the driver's fixation point from another point of view. For the operation of probability, the most intuitive method is to multiply with the input, which is to add this probability information into our feature map.

%The incremental knowledge $\textbf{K}_i$ conveys rich historically-accumulative and long-temporal information. 
%To use $\textbf{K}_i$ to guide the transition of $\textbf{C}_5$ and $\textbf{T}_4$, we first apply a softmax operation on $\textbf{K}_i$ to obtain an attention feature $\textbf{K}_i^{a}$,
To utilize $\textbf{K}_i$ for guiding the transition of $\textbf{C}_5$ and $\textbf{T}_4$, we start by applying a softmax operation on $\textbf{K}_i$ to obtain an attention feature $\textbf{K}_i^{a}$.
\begin{equation}
\textbf{K}_i^{a} = Softmax (\textbf{K}_i),\\
% \label{eq:K_i}
\end{equation}
and then impose $\textbf{K}_i^{a}$ on $\textbf{C}_5$ and $\textbf{T}_4$ by the multiplication:
\begin{equation}
\textbf{T}_4^{g} = \textbf{K}_i^{a} \times {\mathop{\rm LN}\nolimits} ({\mathop{\rm Flatten}\nolimits} (\mathcal{F}_{up}({\textbf{T}_4})))
\label{eq:T4}
\end{equation}
\begin{equation}
\textbf{C}_5^{g} = \textbf{K}_i^{a} \times {\mathop{\rm LN}\nolimits} ({\mathop{\rm Flatten}\nolimits} (\mathcal{F}_{up}({\textbf{C}_5})))
\label{eq:C5}
\end{equation}
where $\mathcal{F}_{up}$ denotes up-sample operation to make $\textbf{T}_4$ and $\textbf{C}_5$ have the same size with $\textbf{K}_i$, LN denotes the layer normalization. By Eqs.~\ref{eq:T4}-\ref{eq:C5}, $\textbf{T}_4$ and $\textbf{C}_5$ are updated under the guidance of $\textbf{K}_i$. $\textbf{T}_4^{g}$ and $\textbf{C}_5^{g}$ are updated Transformer feature and updated CNN feature, respectively.

% As mentioned above, the obtained $\textbf{K}_i$ times $\textbf{C}_{5}$ and $\textbf{T}_{4}$. In order for $\textbf{K}_i$ to be activated as a probability value, hence softmax is applied to it. Inspired by the Attention mechanism of Transformer, the information of $\textbf{C}_{5}$ and $\textbf{T}_{4}$ itself belongs to the same information obtained from two perspectives. This information is essentially the same task, so it is reasonable to conduct an Attention operation on it. For the fusion, the multiplication of $K$ has been applied to their flatten results $f_C$ and $f_T$, and the obtained $Q_C V_C K_T$ matrices are respectively come from $f_C^K$ and $f_T^K$ operated by Linear, which contains the information of $K$. This process can be described as:

% \begin{equation}
% f_T^K = {\textbf{K}_i} \times {\mathop{\rm LN}\nolimits} ({\mathop{\rm flatten}\nolimits} (({T_4})))
% \label{ftk}
% \end{equation}
% The fusion operation can be expressed as:
% \begin{equation}
% Y = {\mathop{\rm softmax}\nolimits} (M_C^q \times M{_T^k{^\top}})
% \label{fusion1}
% \end{equation}
% \begin{equation}
% Z = Y \times M{_C^v}
% \label{fusion2}
% \end{equation}
Given the updated $\textbf{T}_4^{g}$ and  $\textbf{C}_5^{g}$, Transformer is utilized to explore the relationship between them. First, the attention weight is computed by:
\begin{equation}
 \footnotesize
\textbf{W} = \text{Softmax} \left( \frac{{\left( {\textbf{W}_q \textbf{C}_5^{g}} \right) \left( {\textbf{W}_k \textbf{T}_4^{g}} \right)^\mathrm{T}}}{{\sqrt D }} \right)
\label{eq:transformer_k_q}
\end{equation}
where ${ \textbf{C}_5^{g}}$ is \textit{query}, ${ \textbf{T}_4^{g}}$ is \textit{key}, $\textbf{W}_q$ and $\textbf{W}_k$  represent learnable parameters of updating \textit{query} and \textit{key} , and $D$ is the dimension of \textit{key}. 
%$\textbf{W}$ actually indicates the inter-relationship of $\textbf{T}_4$ and $\textbf{C}_5$. 

Then, the fusion feature $\textbf{F}$ is computed by aggregating \textit{Value} (\textbf{V}) weighted with attention weight $\textbf{W}$ in Eq.~\ref{eq:transformer_k_q}:
\begin{equation}
\setlength{\belowdisplayskip}{5pt}
% \footnotesize
\textbf{F} = \textbf{W}\textbf{V}
\label{eq:fusion_feature}
\end{equation}
where $\textbf{V}=\textbf{C}_5^{g}$.
%where we note that \textbf{V} has two configurations, namely $\textbf{V}=\textbf{C}_5$ and $\textbf{V}=\textbf{T}_4$. 

Ultimately, to further enrich the fusion feature, the original CNN feature and the Transformer feature are utilized once again:  
\begin{equation}
   \textbf{F} = \mathcal{F}_{cnn}( \textbf{F} + \mathcal{F}{cnn}(\textbf{C}_{5}) + \mathcal{F}{cnn}(\textbf{T}_{4}))
\label{eq:fusion_feature_update}
\end{equation}
where ${F}_{cnn}(\cdot)$ represents a small CNN block,  ${F}_{cnn}(\cdot) = {\mathop{\rm ReLU}\nolimits} ({\mathop{\rm BN}\nolimits} ({\mathop{\rm Conv}\nolimits}(\cdot)))$. By Eq.~\ref{eq:fusion_feature_update}, the fusion feature is updated, which has considered not only the individual information of $\textbf{T}_4$ and $\textbf{C}_5$ but also the relationship between them.  

% Query, key and value are three inputs of Transformer. We define $\textbf{C}_5$ as the query and $\textbf{T}_4$ as the key.

%  that implies the relationship of $\textbf{T}_4$ and $\textbf{C}_5$

% \[{f_{trans}} = soft\max \left( {\frac{{\left( {{{\rm{W}}_q}{C_5}} \right){{\left( {{{\rm{W}}_q}{T_4}} \right)}^T}}}{{\sqrt d }}} \right) \times \left( {{W_v}{C_5}} \right)\]
% \end{equation}
% \begin{equation}
% f = {\mathop{\rm ReLU}\nolimits} ({\mathop{\rm BN}\nolimits} {\mathop{\rm Conv}\nolimits} (f_{trans} + {C_5} + {\textbf{T}_{4})
% \label{fusion3}
% \end{equation}

% \begin{equation}
%     \mathcal{F}(x) = {\mathop{\rm ReLU}\nolimits} ({\mathop{\rm BN}\nolimits} ({\mathop{\rm Conv}\nolimits}(x)))
% \label{eq:CBR}
% \end{equation}Figure 3: Examples of qualitative comparison. From left to right: image, GT, FBLNet, BDDA, DADA, DRIVE and PGNET.
% In order to make the size of $\textbf{C}_{5}$, $\textbf{T}_{4}$ and $K$ consistent, here we choose to down-sample $K$ and up-sample $\textbf{C}_{5}$, $\textbf{T}_{4}$ at the same time, which also aims at not losing too much image information. So the output fusion feature $\textbf{B}$ size is ${256}\times{14}\times{14}$.
\subsection{Decoder}
\label{section_decoder}
% The inputs of our decoder module are fusion feature ($\textbf{B}$) in Eq.~\ref{eq:fusion_feature}. To strengthen decoder's feature expression $\textbf{C}_i$ in \ref{eq:cnn feature} and $\textbf{T}_i$ in \ref{eq:transformer feature} are added to use skip connection.
The goal of the decoder module is to upsample the fusion feature $\textbf{F}$ to finally predict the driver attention heatmap $\textbf{A}$. The decoder module consists of five blocks denoted as $\{{D}_{i}\}_{i=0}^{4}$, and their corresponding outputs are represented by $\{\textbf{d}_{i}\}_{i=0}^{4}$. 

To make clear each decoder block, let's take ${D}_1$ as an example. The direct input of ${D}_1$ is the output $\textbf{d}_0$ of the previous decoder block ${D}_0$. Besides $\textbf{d}_0$, to further strengthen the expression of the decoder feature, a skip connection mechanism is adapted. Specifically, the CNN feature $\textbf{C}_4$ and Transformer feature $\textbf{T}_3$ skip from the encoder module to the decoder module to serve as the inputs of ${D}_1$. Therefore, $\textbf{d}_0$, $\textbf{C}_4$ and $\textbf{T}_3$ are the inputs of ${D}_1$, which outputs $\textbf{d}_1$. This procedure is simplistically formulated as follows:
\begin{equation}
    {\textbf{d}_1} =  {\mathcal{F}_{up}}\left({\mathcal{F}_{cnn}(Concat(\textbf{C}_{4},\textbf{T}_{3}))} \right) + {\mathcal{F}_{up}}(\mathcal{F}_{cnn}\left( \textbf{d}_{0}\right))
\label{eq:decoder3}
\end{equation}
where $\mathcal{F}_{up}$ denotes up-sample operation, $Concat$ represents the concatenation operation, and ${F}_{cnn}(\cdot) = {\mathop{\rm ReLU}\nolimits} ({\mathop{\rm BN}\nolimits} ({\mathop{\rm Conv}\nolimits}(\cdot)))$ represents a small CNN block.

The output of last decoder block is $\textbf{d}_4$, based on which the driver attention heatmap $\textbf{A}$ is computed:
\begin{equation}
    \textbf{A} = sigmoid\left({Conv({\textbf{d}_4})} \right)
\label{eq:A_output}
\end{equation}

% $\textbf{C}_4$ and $\textbf{T}_3$ from CNN-Transformer Encoder. With this Idea of design, we can deduce $\textbf{d}_2$ and $\textbf{d}_3$.

% The process of $\textbf{d}_1$ block is formulate Figure 3: Examples of qualitative comparison. From left to right: image, GT, FBLNet, BDDA, DADA, DRIVE and PGNET.as:
%Among them, $\textbf{d}_2$ is the special one that we choose to make it feedback feature $\textbf{B}$ (Eq.~\ref{eq:feedback_feature}). For all the other blocks except the first one ($\textbf{d}_0$), we performed layer-hopping join operations to better transform the feature information into the final prediction. This process can be expressed as:
% \begin{equation}
%     {d_i} = \mathcal{F}_{d}\left( {{d_{i - 1}},{\textbf{C}_{5 - i}},{\textbf{T}_{4 - i}}} \right),i \ge 1
% \label{eq:decoder}
% \end{equation}
% \begin{equation}
%     {f_1} = {\mathcal{F}_{up}}\left({\mathcal{F}(Concat(\textbf{C}_{4},\textbf{T}_{3}))} \right)
% \label{eq:decoder1}
% \end{equation}
% \begin{equation}
%     {f_2} = {\mathcal{F}_{up}}\left( {\mathcal{F}\left( {{d_{0}}} \right)} \right)
% \label{eq:decoder2}
% \end{equation}

\section{Experiments}

\subsection{Experimental Settings}

\begin{table*}[!h]
    \footnotesize
    \setlength{\abovecaptionskip}{-0.3cm}
	\caption{Quantitative comparison results on DADA\cite{dada} and BDDA\cite{bdda}  datasets.} 
	\begin{center} 
		\setlength{\tabcolsep}{1.8mm}{
			\begin{tabular}{lccccccccccccc} 
				\toprule
				\multirow{2}*{\textbf{Method}} &\multicolumn{6}{c}{BDDA} &\multicolumn{6}{c}{DADA} \\
				\cmidrule(r){2-7} \cmidrule(r){8-13}
				&AUC\_J$\uparrow$ &AUC\_B$\uparrow$ &SIM$\uparrow$ &CC$\uparrow$ &Kldiv$\downarrow$ &NSS$\uparrow$ 
				&AUC\_J$\uparrow$ &AUC\_B$\uparrow$ &SIM$\uparrow$ &CC$\uparrow$ &Kldiv$\downarrow$ &NSS$\uparrow$ \\
				\hline\hline
				BDDA~\cite{bdda}\tiny{ACCV'2018}  &0.9383 &0.8847 &0.3513 &0.4805 &2.0723 &3.4515
									  &0.8960 &0.8297 &0.2232 &0.3261 &2.7671 &2.5229 \\
				
				U2NET~\cite{U2-Net}\tiny{PR'2020} &{\color{blue}0.9594} &{\color{red}0.9403} &0.3612 &0.5586 &{\color{blue}1.4737} &3.9504
									&0.9481 &{\color{blue}0.9088} &0.3008 &0.4737 &{\color{red}1.8568} &3.7740 \\

				MINET~\cite{pang2020multi}\tiny{CVPR'2020} &0.8645 &0.7546 &0.3571 &0.4882 &10.5031 &4.2808
									  &0.8651 &0.7152 &0.3000 &0.3855 &9.9914 &3.6239 \\
                    DRIVE~\cite{DRIVE}\tiny{ICCV'2021} &0.7637 &0.6890 &0.2589 &0.3158 &13.8349 &2.5615
									  &0.9050 &0.8533 &0.2468 &0.3743 &4.0270 &3.0614 \\
				
				DADA~\cite{dada}\tiny{T-ITS'2021}  &0.9546 &0.9056 &0.3986 &0.5566 &1.4824 &4.2205 
				      				   &{\color{blue}0.9500} &0.8737 &{\color{blue}0.3412} &{\color{blue}0.4797} &2.1689 &{\color{blue}4.0549} \\
				      
				DBNET~\cite{rainy}\tiny{IEEE/CAA'2022} &0.9554 &0.9270 &0.3825 &0.5589 &1.8533 &3.9321
				      					  &0.9190 &0.8792 &0.2582 &0.3957 &2.7716 &2.9301 \\
				
				PGNET~\cite{PGNet}\tiny{CVPR'2022} &0.9237 &0.8184 &{\color{blue}0.4396} &{\color{blue}0.5698} &6.0989 &{\color{blue}4.9290}
			          			      &0.9250 &0.8124 &{\color{red}0.3728} &0.4571 &5.2839 &4.0455\\
			    \midrule
			    \textbf{Ours} &{\color{red}0.9587} &{\color{blue}0.9290} &{\color{red}0.4666} &{\color{red}0.6447} &{\color{red}1.4010} &{\color{red}5.0273} 
			                  &{\color{red}0.9540} &{\color{red}0.9098} &0.3307 &{\color{red}0.5020} &{\color{blue}1.9205} &{\color{red}4.1316} \\
				\bottomrule
				\vspace{-1.0cm}
			\end{tabular} 
		} 
	\end{center}
	\label{tab:compare} 
\end{table*}

\noindent{\textit{\textbf{Datasets}}}. 
% Three public well-known driver attention datasets are selected to conduct experiments.
%{DR(Eye)VE}\cite{dr(eve)vy} is a large driver's attention dataset containing 555,000 frames at 1080p/25fps of 74 videos, which are collected using a car roof-mounted camera by eight drivers in diverse weather conditions (sunny, rainy and cloudy) and various driving situations (downtown, countryside, highway and urban). The annotation of the driver attention is obtained with the help of eye tracking glasses. The annotation for each frame is one or multiple gaze fixation points. A continuous fixation map could be obtained from the fixation points by centering on each of them a Gaussian filter. 
% In our experiment, due to the large scale of the dataset, 
% 16 video are selected to evaluate all methods. 
%The training set contains 52291 frames, and the testing set contains 19154 frames. 
BDDA\cite{bdda} (Berkeley DeepDrive Attention) is a driver attention dataset that focuses on critical situations such as occlusion, truncation, and emergency braking. The dataset is composed of 1,232 videos from different weather and lighting conditions. To obtain the annotations, 45 drivers are asked to observe videos, and their eye movements are recorded by the infrared eye tracker to generate attention points. The ground truth attention map of each frame is obtained by smoothing the attention point aggregation of several observers. 
%In this experiment, we choose 15768 frame as the training set, and 29200 frame as the testing set.
% The BDDA dataset, which focuses on drivers' gaze points in crisis situations, is a combination of 1,232 videos from different weather and lighting conditions. Each video is short in length, only about ten seconds. Each frame is the average result marked by multiple experimenter, and its video picture content is richer, including a variety of occlusion, truncation and emergency braking, and the average number of people and vehicles in each frame is also larger.Since crisis situations are rarely encountered in real driving situations, the collection of driver gaze points in this dataset is not in the vehicle, but in a safe environment. In this experiment, we choose xxx frame as the training set and xxx frame as the test set.
DADA\cite{dada} is a driver attention dataset involving 54 kinds of accident scenarios, including 658,476 video frames of 2,000 videos collected in various scenes (highways, urban, rural, and tunnel), weather conditions and light conditions. 
% The dataset is well organized according to the accident types.
With the help of the infrared eye tracker, 20 volunteers annotated multiple attention points for each frame, and the ground truth attention map is obtained by applying Gaussian filters on attention points. 
% Due to the relatively large scale of dataset, a part of data are selected in the experiments, and the training set containing 13301 frame, testing set contains 18631 frame.

% The driving environment of the DADA dataset is somewhat similar to that of BDDA, which is for crisis situations. But the difference is that he videos in the DADA data set are of incidents that have already happened, which contains 658476 video frame. DADA includes 2000 in various occasions (highways, urban, rural and tunnel), weather, sunny, rain and snow) and light conditions (day and night) driving video, they are divided into 54 kinds of video of traffic accident happened. Each frame is annotated by multiple experimenters, and classifies video according to the accident type. Due to the relatively large data sets, our study chose to use the previous six kind of accident video as the training set and test set, the training set containing XXX frame, frame test set contains XXX.

\noindent{\textit{\textbf{Evaluation Metrics}}}.  
Six classical metrics, including three distribution-based metrics: Pearson's Correlation Coefficient (CC), Kullback-Leibler divergence (Kldiv), and Similarity (SIM). Three location-based metrics: Normalized Scanpath Saliency (NSS) and two types of  Area Under the ROC curve (AUC\_J\cite{aucj}, AUC\_B\cite{aucb}) are selected for the performance evaluation.
AUC is to calculate the Area Under the ROC which is created by plotting the true positive rate (TPR) against the false positive rate (FPR).
\begin{equation}
    \footnotesize
    FPR = \frac{{FP}}{{FP + TN}},TPR = \frac{{TP}}{{TP + FN}}
\end{equation}
  CC calculates the linear relationship of the prediction ($\textbf{P}$) and ground truth ($\textbf{Q}$).
\begin{equation}
    \footnotesize
    CC(\textbf{P},\textbf{Q}) = \frac{{Cov(\textbf{P},\textbf{Q})}}{{\sigma (\textbf{P}) \times \sigma (\textbf{Q})}}
\end{equation}
Kldiv  is used to measure the difference between two probability distributions.
\begin{equation}
    \footnotesize
Kldiv(\textbf{P},\textbf{Q}) = \sum\nolimits_{i}{\textbf{Q}_i log\left( \varepsilon + \frac{\textbf{Q}_{i}}{\varepsilon + \textbf{P}_{i}}  \right)} 
\end{equation}
SIM indicates the similarity between $P$ and $Q$. 
\begin{equation}
    \footnotesize
SIM\left( {\textbf{P},\textbf{Q}} \right) = \sum\nolimits_{i}{min\left( \textbf{P}_{i},\textbf{Q}_{i}  \right)} 
\end{equation}
NSS is used to evaluate the difference between $P$ and $Q$. 
\begin{equation}
    \footnotesize
    NSS\left( {\textbf{P},\textbf{Q}} \right) = \frac{1}{N}\sum\nolimits_i {\left( {\overline {{\textbf{P}_i}}  \times {\textbf{Q}_i}} \right)} 
\end{equation}
For AUC, CC, SIM and NSS metrics, higher values are expected. For the Kldiv metric, a lower value is expected. 

\noindent{\textit{\textbf{Implementation Details}}}.
% For our decoder structure, ResNet-18 is used as CNN branch and for the Transformer brach we use Swin-Transformer. So, the input image needs to be resized to ${3}\times{224}\times{224}$ to conform the requirement of Swin-Transformer. The Adam optimizer is applied with the initial learning rate set to $10^{-4}$ and learning rate is reduced to 0 for a specified number of epochs, each of which is reduced by the same amount. Momentun and weight decay are set to 0.9 and $10^{-4}$, respectively. The batch size of training is 16. And the model implementation is based on PyTorch.
% The Encoder of FBL is contained two branch, a CNN branch for extracting local features and a Transformer branch focusing on global features,
\textbf{1)} The backbone of our CNN branch in Eq.~\ref{eq:cnn feature} is ResNet-18\cite{resnet18}. \textbf{2)} Swin-Transformer\cite{swinT} is used as the backbone of our Transformer branch in Eq.~\ref{eq:transformer feature}. \textbf{3)} The size of input image $\textbf{I}$ is ${3}\times{224}\times{224}$ to conform the requirement of Swin-Transformer. \textbf{4)} We choose $\textbf{d}_2$ as our feedback feature Eq.~\ref{eq:feedback_feature}. \textbf{5)} The size of incremental knowledge is initially defined as ${64}\times{56}\times{56}$. And it is resized to ${256}\times{14}\times{14}$ before entering into fusion module. \textbf{6)} The number of channels for $\textbf{T}_4$ and $\textbf{C}_5$ are both 512, a squeeze channel operation is applied to reduce the channel number to 256 so that it meets the requirement of the size of $\textbf{B}$. \textbf{7)} The Adam optimizer is used with the initial learning rate set to $10^{-4}$, the momentum and the weight decay are set as 0.9 and $10^{-4}$, respectively. \textbf{8)} The batch size is set to 8, and experiments are conducted in an NVIDIA RTX3090 GPU. \textbf{9)} Our model implementation is based on PyTorch.

\noindent{\textit{\textbf{Loss Function}}}.
The loss function $\mathcal{L} (\textbf{P},\textbf{Q})$ is defined between the predicted saliency map $\textbf{P}$ and the corresponding ground-truth saliency map $\textbf{Q}$. 
\begin{equation}
 \footnotesize
\mathcal{L} (\textbf{P},\textbf{Q}) = \mu Kldiv(\textbf{P},\textbf{Q}) -  \eta NSS(\textbf{P},\textbf{Q})  - \xi CC(\textbf{P},\textbf{Q}) 
\end{equation}
where $\mu = 1$, $\eta = 0.1$, and $\xi = 0.1$ are the scaler factors for Kldiv, NSS, and CC, respectively. 
\subsection{Comparison Experiment}

\begin{figure*}[!t]
	\centering
	\setlength{\abovecaptionskip}{0.cm}
      \setlength{\belowcaptionskip}{-0.3cm}
	\includegraphics[width=1\textwidth]{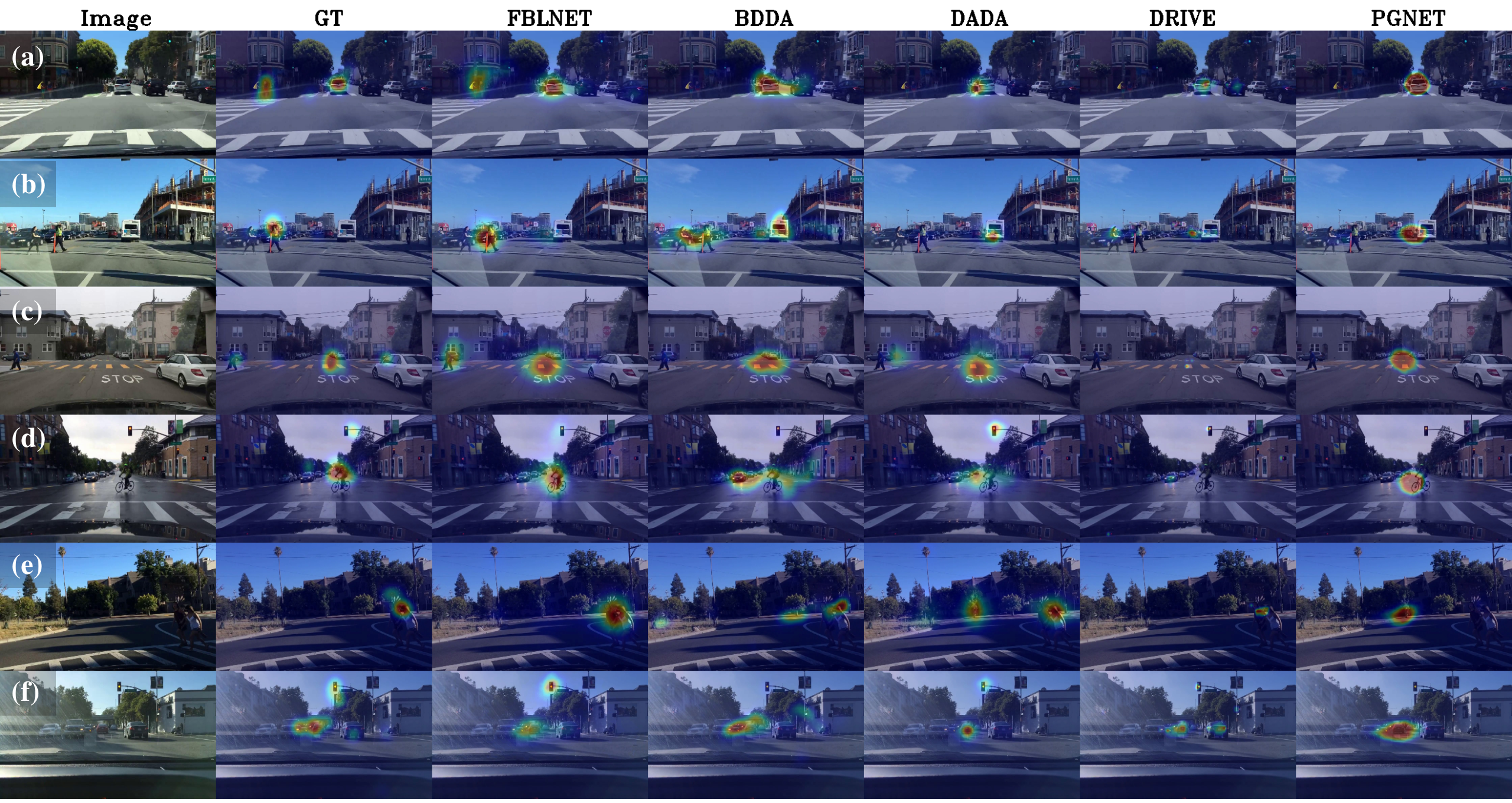}

 \caption{Qualitative comparison of our method (i.e,  FBLNet) with baselines (i.e., BDDA\cite{bdda}, DADA\cite{dada}, DRIVE\cite{DRIVE} and PGNet\cite{PGNet}).}
	\label{fig_compare}
\end{figure*}
\noindent{\textit{\textbf{Baselines}}}. 
In order to illustrate the performance of our model, we compared it with other seven models, among which DBNet\cite{rainy}, DRIVE\cite{DRIVE}, U2Net\cite{U2-Net},  MINet\cite{pang2020multi}, BDDA\cite{bdda} and DADA\cite{dada} are completely based on CNN, and PGNet\cite{PGNet} is based on CNN and Transformer. 
%All models are retrained on the dataset we selected, and the best parameters and results are preserved.

\noindent{\textit{\textbf{Quantitative Comparison}}}. 
Tab.~\ref{tab:compare} shows the comparison results on BDDA\cite{bdda} and DADA\cite{dada} datasets.
We can observe that our model overall outperforms all seven baselines. Compared with the second-best result on SIM, CC and Kldiv metrics, our model achieves  6.14\%, 13.4\%, and 7.27\% performance improvement, respectively, on the BDDA dataset.  On the DADA dataset, our model achieves 4.65\% and 3.43\% performance improvement, on CC and Kldiv metrics, respectively.
%our model achieves the average 9.09\%, 1.62\%, and 20.3\% performance improvement on three datasets, respectively. 
Especially, solid comparison margins are obtained on the BDDA dataset across CC and Kldiv metrics. The  BDDA dataset focuses on accident scenarios where driver attention prediction is extremely important for danger warnings. 

In addition, the comparison results indicate that our proposed model structure presents the advantage over the commonly-used Encoder-Decoder structure (i.e., DBNet \cite{rainy}, DRIVE\cite{DRIVE}, U2Net\cite{U2-Net} and MINet\cite{pang2020multi}) and recently-proposed CNN+Transformer structure (i.e., PGNet\cite{PGNet}). The superiority of our model structure is three-fold: \textbf{1)} A novel FeedBack Loop (FBL) network is proposed. FBL continuously fetches decoder features backward to the incremental knowledge, which is able to guide driver attention prediction using historically-accumulative long-range temporal information. \textbf{2)} A incremental knowledge guided and Transformer based fusion network is proposed. This fusion mechanism not only strengthens the CNN feature and Transformer feature by applying the incremental Knowledge, but also digs out the inter-relationship among all features via the Transformer idea.
\textbf{3)} CNN-Transformer encoder is designed to extract features. CNN presents the advantage of extracting detailed local features, and Transformer is powerful at modeling long-range global contexts.

\noindent{\textit{\textbf{Qualitative Comparison}}}. 
Fig.~\ref{fig_compare} illustrates some examples of qualitative comparison results.  We list six complex scenarios and they are represented by (a)-(f).
On the whole, our model outperforms all baselines. Especially in scenes (a), (b) and (c), our model shows great superiority.
%In relatively easy scenarios where the driver has single attention points, all models achieve good performance. When confronting difficult scenarios where the driver should pay attention to multiple points, our model exhibits better performance.
%This is particularly evident in the first two images from top to bottom. 
% 1) In scene (a), it has two unobtrusive but important objects that need attention (i.e., two pedestrians crossing the road) in the left shaded area, which are difficult for the models without FBL to predict, but it is clear that FBLNet is focusing to this zone. 
% 2) In scene (b), no baseline method performs as well as ours, they all predict the white vehicle with the most salient color as the driver attention, while what should really be paid attention to is the pedestrian who is crossing the road on the left.
% 3) In scene (c),  FBLNet notices the pedestrian in the left, while other methods even do not pay attention to the pedestrian.
% 4) In scene (d), it has two salient objects (i.e., a traffic light and a bicyclist), the bicyclist is large and the light is small. 
%  Only FBLNet detects these two objects with different size well.
% 5) In scene (e), three baseline methods ruin the attention area.
1) In scene (a), two pedestrians crossing the road in the left shaded area demand attention, posing a challenge for models without FBL. However, FBLNet accurately focuses on this zone.
2) In scene (b), no baseline method matches the performance of FBLNet, as they all wrongly predict the white vehicle with the most salient color as the driver's attention, while the actual focus should be on the pedestrian crossing on the left.
3) In scene (c), FBLNet correctly notices the pedestrian on the left, while other methods neglect this important object.
4) In scene (d), with a large bicyclist and a small traffic light, only FBLNet effectively detects both objects of different sizes.
5) In scene (e), three baseline methods fail to accurately identify the attention area, while FBLNet stands out with superior results.

\subsection{Diagnostic Experiment}
\label{subsection:Diagnostic Experiment}
To examine the detailed impacts of crucial components (i.e., \textbf{Feedback Loop},  
%\textbf{Residual- and Transformer-Inspired Fusion}, 
and \textbf{CNN-Transformer Encoder}) in our model, a set of ablative studies are conducted on DADA\cite{dada} and BDDA\cite{bdda} datasets.

\begin{figure}[t]
	\centering
	\setlength{\abovecaptionskip}{0.cm}
      \setlength{\belowcaptionskip}{-0.2cm}
	\includegraphics[width=0.475\textwidth]{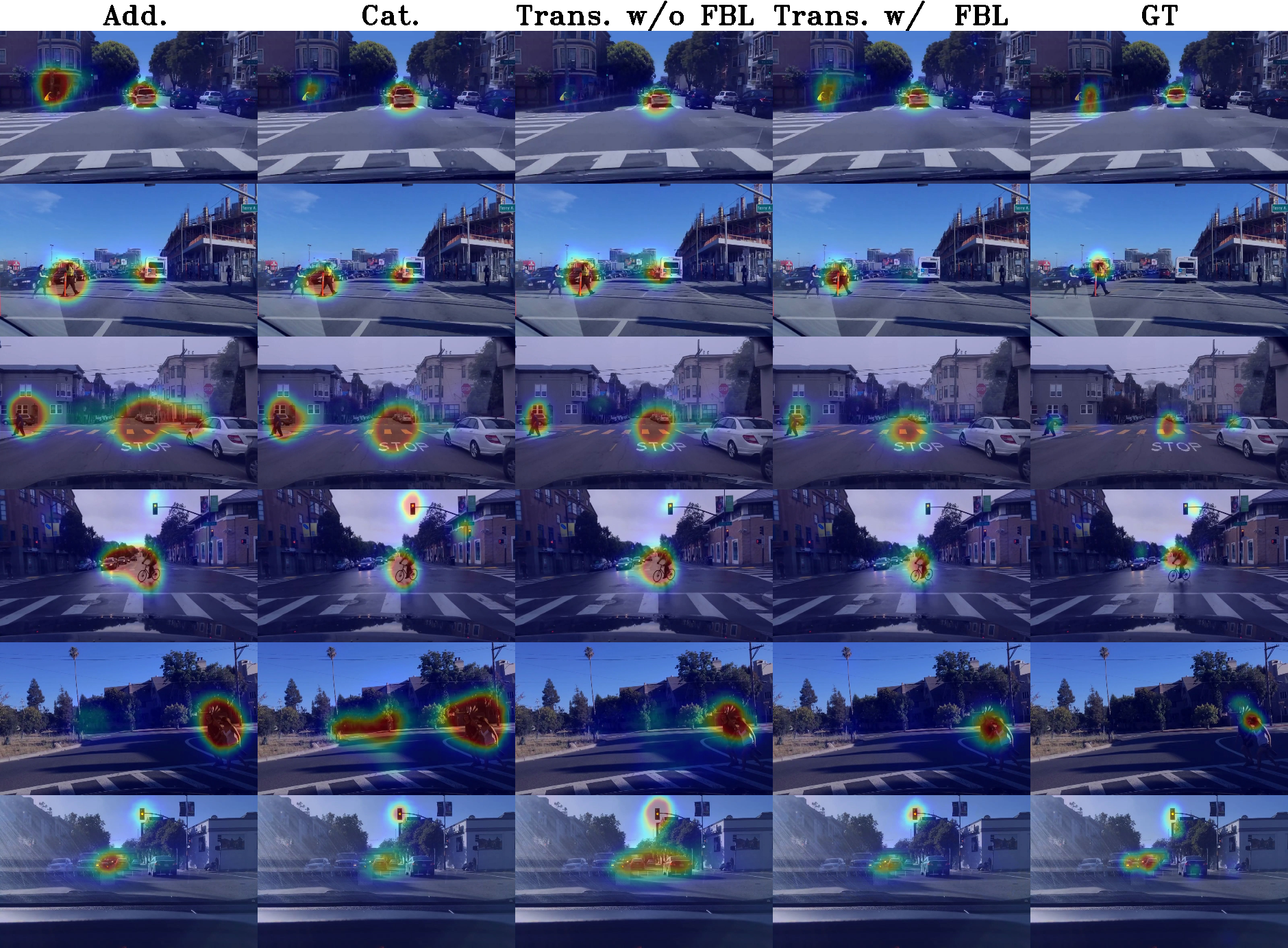}
 \caption{Visualization on the ablation study on our FBL. We can clearly see that the comparison methods present a divergence in the prediction region, inaccurate prediction and even wrong prediction in these complex scenarios}
    \label{fig_fusion}
 \end{figure}

 \begin{table}[t]
	\footnotesize
	\setlength{\abovecaptionskip}{0.3cm}
	\caption{Ablation study on different types of FBL.}
      \vspace{-0.1cm}
	\begin{center} 
		\setlength{\tabcolsep}{1.2mm}{
			\begin{tabular}{lccccccc} 
				\toprule
				\multirow{2}*{\textbf{Fusion}} &\multicolumn{3}{c}{DADA} &\multicolumn{3}{c}{BDDA} \\
				\cmidrule(r){2-4} \cmidrule(r){5-7}
				&SIM$\uparrow$ &CC$\uparrow$ &NSS$\uparrow$ &SIM$\uparrow$ &CC$\uparrow$ &NSS$\uparrow$ &avg$\uparrow$\\
				\hline\hline 
				Cat. &0.3087 &0.4721 &3.6621 &0.4358 &0.6134 &4.5741 &1.6777\\
				Add.   &0.3254 &0.4869 &3.9035 &0.4416 &0.6116 &4.6454 &1.7357\\
				%Trans. $(V=T_4)$ &0.3152 &0.4847 &3.7801 &0.4611 &0.6394 &4.8579  &1.7564\\
                   w/o FBL  &0.3204 &0.4864 &3.8658 &0.4561 &0.6352 &4.8102 &1.7624\\
				w/ FBL &{\textbf{0.3307}} &{\textbf{0.5020}} &{\textbf{4.1316}} &{\textbf{0.4666}}  &{\textbf{0.6447}} &{\textbf{5.0273}} &{\textbf{1.8505}}\\
                    %Cat. w/o fb &0.3120 &0.4794 &3.6898 &0.4560 &0.6338 &4.7881 &1.7265\\
                    %Add. w/o fb &0.3212 &0.4852 &3.8404 &0.4565 &0.6338 &4.8082 &1.7576\\
				\bottomrule
				\vspace{-1.cm}
			\end{tabular} 
		} 
	\end{center}
	\label{tab:fusion} 
\end{table}

\noindent{\textit{\textbf{Feedback Loop}}}. First, we investigate the essential feedback loop network, and the results are shown in Tab.~\ref{tab:fusion}.  From the table we can see the performance under different fusion mechanisms on the BDDA\cite{bdda} and DADA\cite{dada} datasets.
In the table, Add. and Cat. are two simple ways (i.e., addition and concatenation) we used to fuse the different features from the Transformer path and the CNN path. 
w/o FBL and w/ FBL represent our incremental knowledge guided fusion module with FBL and without FBL, respectively.
An average evaluation (i.e., avg) is adopted to show the overall performance of different methods. 
Compared with the simple fusion methods, our incremental knowledge guided fusion method exhibits the advantage, averagely obtaining 6.61\% and 10.30\% performance higher than the addition fusion method and concatenation fusion method, respectively.
Meanwhile, compared with the w/o FBL, the method w/ FBL gains performance improvement on all metrics. 
The results are an average of 6.23\% higher than that of configuration without FBL. 
The reason is explainable. 
When FBL is disabled, the incremental knowledge $\textbf{K}_i$ (Eq.~\ref{eq:K_i}) is also disabled, thus the information carried in $\textbf{K}_i$ could not guide the prediction, and simple fusion methods do not integrate features properly, therefore leading to the performance degradation. By fusing $\textbf{K}_i$ with the CNN feature and the Transformer feature, the model makes full use of the historically-accumulative information to predict driver attention, thus bringing performance improvement. These experiments validate the effectiveness of FBL, and prove that $\textbf{K}_i$ plays a crucial role in driver attention prediction.
 They also prove that Transformer mechanism is helpful for performance improvement since it allows the network to learn the potential inter-relationship between different features.
%Furthermore, with the help of FBL, the fusion is better, indicating that the FBL further strengthens the ability to learn this potential inter-relationship.

To further explore the significance of FBL, we have done the visualization of the results of this experiment. 
%From the Fig~\ref{fig_fusion}, we can clearly see that when we use FBL, the model's prediction results are closer to GT, while other methods present an inaccurate prediction and even wrong prediction. The diffusion phenomenon of the concatenation method and addition method is obvious, meanwhile the incremental knowledge guided fusion method without FBL does not have a complete prediction.
From Fig.~\ref{fig_fusion}, it is evident that when FBL is employed, the model's prediction results are significantly closer to GT compared to other methods. The concatenation and addition methods show apparent diffusion phenomena, resulting in inaccurate and even incorrect predictions. On the other hand, the incremental knowledge-guided fusion method without FBL fails to achieve a complete prediction. The superiority of FBL in guiding the fusion process is evident from the improved accuracy and reliability of the model's predictions.

 \begin{table}[t]
	\footnotesize
	\setlength{\abovecaptionskip}{0cm}
      \setlength{\belowcaptionskip}{0cm}
	\caption{Ablation study on feedback node.} 
	\begin{center} 
		\setlength{\tabcolsep}{1.2mm}{
			\begin{tabular}{lccccccc} 
				\toprule
				\multirow{2}*{\textbf{Method}} &\multicolumn{3}{c}{DADA} &\multicolumn{3}{c}{BDDA} \\
				\cmidrule(r){2-4} \cmidrule(r){5-7}
				&SIM$\uparrow$ &CC$\uparrow$ &NSS$\uparrow$ 
				&SIM$\uparrow$ &CC$\uparrow$ &NSS$\uparrow$  &avg$\uparrow$\\
				\hline\hline
				%$\textbf{B}=\times$ &0.3204 &0.4864 &3.8658 &0.4561 &0.6352 &4.8102 &1.7624\\
				$\textbf{B}=\textbf{d}_0$ &0.3244 &0.4901 &3.9441 &0.4492 &0.6320 &4.7351 &1.7625\\
				$\textbf{B}=\textbf{d}_1$ &0.3232 &0.4921 &3.9253 &0.4649 &0.6380 &4.8873 &1.7885 \\
				$\textbf{B}=\textbf{d}_2$ &{\textbf{0.3307}} &{\textbf{0.5020}} &{\textbf{4.1316}} &{\textbf{0.4666}}  &{\textbf{0.6447}} &{\textbf{5.0273}} &{\textbf{1.8505}}\\ 
				$\textbf{B}=\textbf{d}_3$ &0.3303 &0.4888 &3.9432 &0.4616 &0.6344 &4.8526  &1.7852\\
				$\textbf{B}=\textbf{d}_4$ &0.3219 &0.4846 &3.8470 &0.4615 &0.6384 &4.8539 &1.7679\\
				\bottomrule
			\end{tabular} 
		} 
	\end{center}
      \vspace{-0.8cm}
	\label{tab:feedbackloop} 
\end{table}
 Having demonstrated the effectiveness of the feedback loop, It is also important to diagnose ``where to start feedback could achieve optimal performance''.
In Tab.~\ref{tab:feedbackloop}, $\textbf{B} = \textbf{d}_i$ denote the configuration with the feedback from $\textbf{d}_i$.
%Having demonstrated the effectiveness of the feedback loop, it is also interesting to diagnose ``which kind of feedback loop could achieve optimal performance'' and ``what is the potential reason behind it''.
Intuitively, the configuration $\textbf{B}=\textbf{d}_4$ (Eq.~\ref{eq:feedback_feature}) should be the optimal solution since it is in line with our Wiener's cybernetics inspired design motivation that uses the feedback of output to make the model achieve better performance. However, 
as shown in Tab.~\ref{tab:feedbackloop}, the model presents the best overall performance when $\textbf{B}=\textbf{d}_2$. The potential reasons are manifold. First, the global information in decoder feature $\textbf{d}_i$ gradually weakens while local information gradually strengthens with the increasing of decoder layers, and $\textbf{B}=\textbf{d}_2$ is a trade-off. Second, if the early decoder features (i.e., $\textbf{d}_0$ and $\textbf{d}_1$) are selected as the feedback feature, the feedback node is excessively near to the fusion node, making the feedback meaningless. Third, if the late decoder feature (i.e., $\textbf{d}_3$ and $\textbf{d}_4$) is selected as the feedback feature, there exists a large gap among the sizes of the feedback feature, CNN feature, and Transformer feature. Therefore, a necessary down-sampling operation is needed, which leads to the loss of key information.

\begin{table}[t]
	\setlength{\abovecaptionskip}{0cm}
	\caption{Ablation study on encoder module.}
      \footnotesize
	\begin{center} 
		\setlength{\tabcolsep}{0.85mm}{
			\begin{tabular}{lccccccc} 
				\toprule
				\multirow{2}*{\textbf{Encoder}} &\multicolumn{3}{c}{DADA} &\multicolumn{3}{c}{BDDA} \\
				\cmidrule(r){2-4} \cmidrule(r){5-7}
				&SIM$\uparrow$ &CC$\uparrow$ &NSS$\uparrow$ &SIM$\uparrow$ &CC$\uparrow$ &NSS$\uparrow$ &avg$\uparrow$\\
				\hline\hline 
				CNN   &0.3112 &0.4832 &3.7695 &0.4550 &0.6376 &4.8660 &1.7538 \\
				Trans.   &0.3117 &0.4805 &3.7377  &0.4513 &0.6327 &4.7451 &1.7265\\ 
				CNN + Trans. 	 &{\textbf{0.3307}}  &{\textbf{0.5020}}  &{\textbf{4.1316}} &{\textbf{0.4666}}  &{\textbf{0.6447}}  &{\textbf{5.0273}}  &{\textbf{1.8505}} \\
				\bottomrule
			\end{tabular} 
		} 
	\end{center}
	\vspace{-0.8cm}
	\label{tab:encoder} 
\end{table}

\noindent{\textit{\textbf{CNN-Transformer Encoder}}}.
We analyze the performance of different encoder types, and the results are summarized in Tab.~\ref{tab:encoder}. In the table, three encoder types are compared, and we can observe that the encoder combining both CNN and Transformer generates impressive results, achieving 5.15\% and 7.18\% average performance improvement compared with the only CNN-based and Transformer-based encoders respectively. The reason is  CNN presents the advantage of extracting local features using small size convolution windows but lacks the ability to globally encode long-range dependencies. As a complement, Transformer is powerful at modeling global long-range contexts. 

\noindent{\textit{\textbf{Discussion}}}.
% Therefore, our CNN-Transformer encoder performs better.
%Let's discuss some of the more challenging scenarios which should be devoted to more research in the future. These scenarios are very difficult for all current models, but they are also very common and important in real driving.
There exist several typical scenarios that are challenging for existing methods. However, they are common but very important for real driving, thus should be devoted to more research focuses in the future. 
\textbf{1)} Salient object is not consistent with the real driver attention. 
%For example, when ego car intends to turn right in an intersection, a stilled object with salient color in front of the ego car is often predicted as the driver attention by the saliency-based methods, but it is actually not the real driver attention since the driver pay more attention on the right-turning driving line.  
\textbf{2)} A scene does not have any salient traffic participant in current time.
% For example, in an intersection with no car or pedestrian, existing visual saliency prediction methods can hardly predict driver attention. However, an experienced driver could pay attention on the zones (e.g., the entrance of a lane that a car could merge) where the danger most likely appear in the short future. 
\textbf{3)}  Many participants share the similar probability to be predicted as driver attention. 

\section{Conclusion}
%discussion
%During our training and testing, we found that there were several scenarios that were very difficult for all models. \textbf{1)} Intersections with fewer objects in the picture, usually occur when waiting at intersections. At this time, there is a lack of obvious vehicles and pedestrians in the picture, which are the most noticed in other scenes. The prediction of such a scenario requires the model to have some kind of ``driving experience" knowledge, so that the model can predict such seemingly insignificant places as the intersection at the edge of the view, where there may be a vehicle coming in. This scene is also very dangerous. Some accidents may be avoided if we can inform the driver to pay attention to these inconspicuous positions in advance. Therefore, solving this question is valuable to improve the driving safety. \textbf{2)} Scenes with many objects in the picture. For example, on a road with a lot of traffic, there are many areas that can be predicted by the model. However, each area also has different importance. For example, the car inserting is more important than the vehicle in front or the pedestrian who is far away. A driver can judge who is more important based on his driving experience, but a machine don't have. So, one way to solve the problem of driver prediction is for machines to own the experience of the driver.

%conclusion
This paper handles the problem of driver attention prediction. Different from the existing mainstream methods, the proposed model presents several novelties. First, FBLNet is proposed to enable the model to simulate the human-like driving experience accumulation. Second, the CNN-Transformer encoder is designed to simultaneously extract local and global features from the input image. Third, a Transformer based fusion mechanism is proposed to fuse the CNN feature and the Transformer feature under the guidance of the incremental knowledge. Some conclusions are drawn through extensive experiments. We list the conclusions here and hope they could benefit the related studies in the community. 1) CNN-Transformer encoder is better than the individual CNN or Transformer encoder. 2) Our proposed feedback loop mechanism is effective, because it provides the incremental knowledge containing the accumulated long-range temporal information like human driving experience.

{\small
\bibliographystyle{ieee_fullname}
\bibliography{final}
}
\end{document}